\documentclass{article}

\usepackage[final]{neurips_2019}

\usepackage[utf8]{inputenc} % allow utf-8 input
\usepackage[T1]{fontenc}    % use 8-bit T1 fonts
\usepackage{hyperref}       % hyperlinks
\usepackage{url}            % simple URL typesetting
\usepackage{booktabs}       % professional-quality tables
\usepackage{amsfonts}       % blackboard math symbols
\usepackage{nicefrac}       % compact symbols for 1/2, etc.
\usepackage{microtype}      % microtypography
\usepackage{graphicx}

\title{On the Similarity of Deep Learning Representations Across Didactic and Adversarial Examples}

\author{%
  PK Douglas\\ 
  Institute for Simulation and Training\\
  Psychiatry and Biobehavioral Medicine, UCLA \\
  \texttt{pkdouglas16@gmail.com} \\
  \And
  Farzad Vasheghani Farahani \\
  Department of Industrial Engineering\\
  University of Central Florida \\
  {farzad.vasheghani@gmail.com}\\
}

\begin{document}

\maketitle

\begin{abstract}
  The increasing use of deep neural networks (DNNs) has motivated a parallel endeavor: the design of adversaries that profit from successful misclassifications.  However, not all adversarial examples are crafted for malicious purposes.  For example, real world systems often contain physical, temporal, and sampling variability across instrumentation. Adversarial examples in the wild may inadvertently prove deleterious for accurate predictive modeling. Conversely, naturally occurring covariance of image features may serve didactic purposes. Here, we studied the stability of deep learning representations for neuroimaging classification across didactic and adversarial conditions characteristic of MRI acquisition variability. We show that representational similarity and performance vary according to the frequency of adversarial examples in the input space.

\end{abstract}

\section{Introduction}

Deep neural networks (DNNs) have transformed computational analysis across a variety of domains from robotic control devices (Abolghasemi 2019) to genomics (Zou et al. 2019) to neuroimaging (Douglas et al. 2013).  Despite their success, DNNs can be susceptible to adversarial examples, or examples that are only slightly different from correctly classified examples and drawn from the same distribution (Goodfellow et al. 2015).  Interestingly, adversarial examples that produce misclassifications in one network will often succeed at producing errors across a range of neural network architectures (Szegedy et al. 2014). 

In some cases adversarial examples are crafted with the intention of deceiving a classifier, potentially posing threats to security and privacy. However, naturally occurring variation such as sunlight damage or graffiti modification can fool traffic signs classifiers (Evtimov et al. 2017). In some cases, adversarial examples are so similar to the original examples that they are imperceivable to the human eye (e.g., Deng et al. 2009). This is often the case when a small amount of bounded noise is applied over the entire input (Tomsett et al. 2018). This suggests that the human visual system may in some way learn the noise more effectively, or be less vulnerable to adversarial perturbations.

Within the field of neuroimaging, however, the goal is often not only to predict accurately but also to interpret which aspects of the image gave rise to a prediction (Kriegeskorte and Douglas 2019).  For example, visualization of hidden layer representations has provided insight into the functional response patterns in the human visual processing stream (Khaligh-Razavi et al 2014), and saliency maps may provide critical information for interpreting a diagnosis.  However, many of these relevance algorithms can be numerically unstable under certain circumstances, and their performance on clinical neuroimaging data remains unclear. Appreciating how adversarial examples not only alter model performance but also effect representations and interpretations is an area in need of further research.

Stochastic rounding, teacher noise, and stochastic shakes introduce noise at varying stages of the training process that can sometimes improve classifier performance (e.g., Gastaldi 2017). However, noise - carefully crafted or otherwise - can also promote mislabeling. We therefore approached the present study free from hypotheses.

\subsection{Related work}

Visualizing what a neural network has learned poses many challenges. Multiple techniques for evaluating support features have recently been developed to provide insight into otherwise black box classifiers. Salience maps may be computed by simply observing output sensitivity to localized input perturbations (Zeiler and Fergus 2014).   Here, we study relevance maps calculated using LIME (Ribeiro et al. 2016), and layer-wise relevance propagation (LRP) (Bach et al. 2015), recently applied to neural network decisions in MRI-based classification of Alzheimer’s disease (Böhle et al. 2019).

\begin{figure}[h!]
  \centering
   \includegraphics[width=12cm]{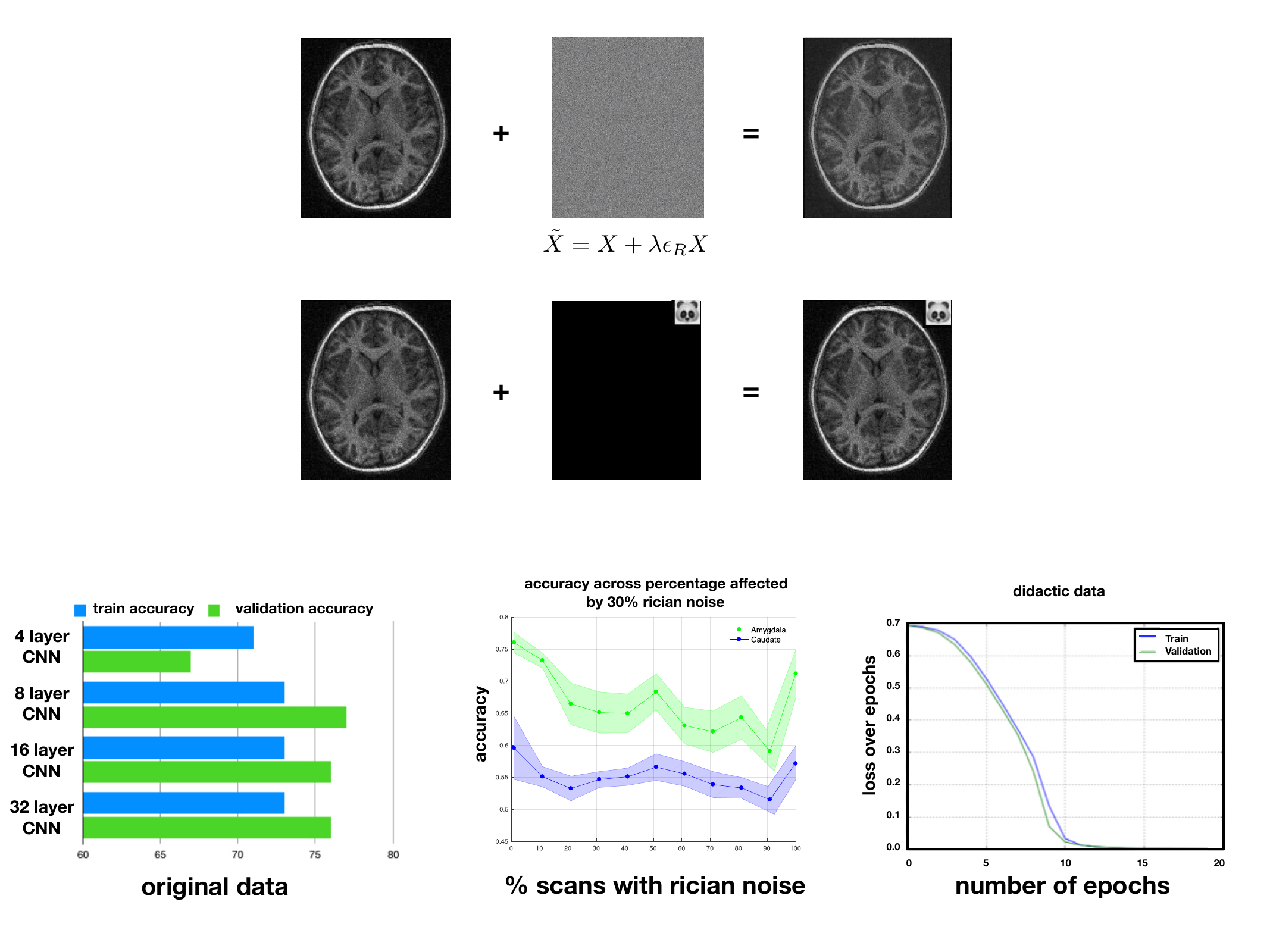}
    \caption{ (Top) Adversarial exemplars $\tilde{X}$ were created with the addition of noise found commonly in the MR setting. (Middle) \emph{Didactic} examples were created by placing an image (a panda) in a location that varied according to the class label. The image was added to the top right non-brain region for ADHD subjects and top left for TD. (Lower from left to right) Accuracy across different CNN models using original data; accuracy when percentage of noise exemplars varies, shown for Rician noise $\lambda$=0.15 ;  Loss over training epochs when all exemplars were didactic.}
    \label{fig:Figure1.001.jpeg}
\end{figure}

\section{Deep Learning with Adversarial and Didactic Examples} 

We used axial slices derived from T1-weighted structural MRI data.  We randomly selected a subset of subjects from the ADHD200 and UCLA TRECC data sets.  In total, there were 1000 scans included.  Of these, 500 subjects were diagnosed with ADHD, while the remaining 500 subjects were typically developing (TD) adolescents.  From these images, we extracted amygdala and caudate slices based on the Harvard Oxford  probability atlas.  Based on previous work, these subcortical structures are expected to carry diagnostic information (Douglas et al. 2018; Hoogman et al. 2017). We found that an 8 layer CNN model trained with stochastic gradient descent achieved levels comparable to the ADHD200 machine learning contest. This model was therefore used for the remainder of the study. 

Bounded noise is the most studied form of \emph{linear} adversarial perturbation. To create an adversarial image, $\tilde{X}$, we corrupted the original image, \emph{X}, with bounded noise $\epsilon$, common in the MR setting, that followed either a Rician $\epsilon_R$, Chi-squared $\epsilon_X$, or Gaussian distribution $\epsilon_G$, that varied according to the fractional variance $\lambda$ of the image intensity. To create \emph{didactic} examples, we added a small image in a non-brain region to instruct the classifier. In this case, we chose to use a small image of a panda, placed in the upper left or right corner of the axial image according to the class label (Figure 1).

\section{Relevance Structural Similarity Analysis}

 Structural similarity is a non-negative index that compares contrast (c), luminance (l), and structural similarity (s) between images (Zhou et al. 2004). Here, we performed relevance structural similarity analysis (RSSA) by extending this method to compare relevance heat maps (r) for images X and their corrupted counter part $\tilde{X}$ as: 
 \begin{equation}
RSSA( \tilde{X_r}, X_r)=[{l(\tilde{X_r},X_r) \cdot c(\tilde{X_r}, X_r) \cdot s(\tilde{X_r}, X_r)}]   
\end{equation}

 The addition of noise (up to $\lambda$ = 0.2) to the \emph{entire} set of data exemplars appeared to have little effect on the classifier performance (Figure 1).  This prompted us to vary the percentage of images that were corrupted, as would be expected if data acquisition took place at different instrumentation sites. Classifier performance across adversarial exemplar frequency, and group wise RSSA maps are shown in Figure 2. In the didactic case, both LIME and LRP assigned a significant amount of relevance to brain regions, suggesting that the prediction did not operate on the didactic signal alone.

\begin{figure}[h!]
  \centering
   \includegraphics[width=12cm]{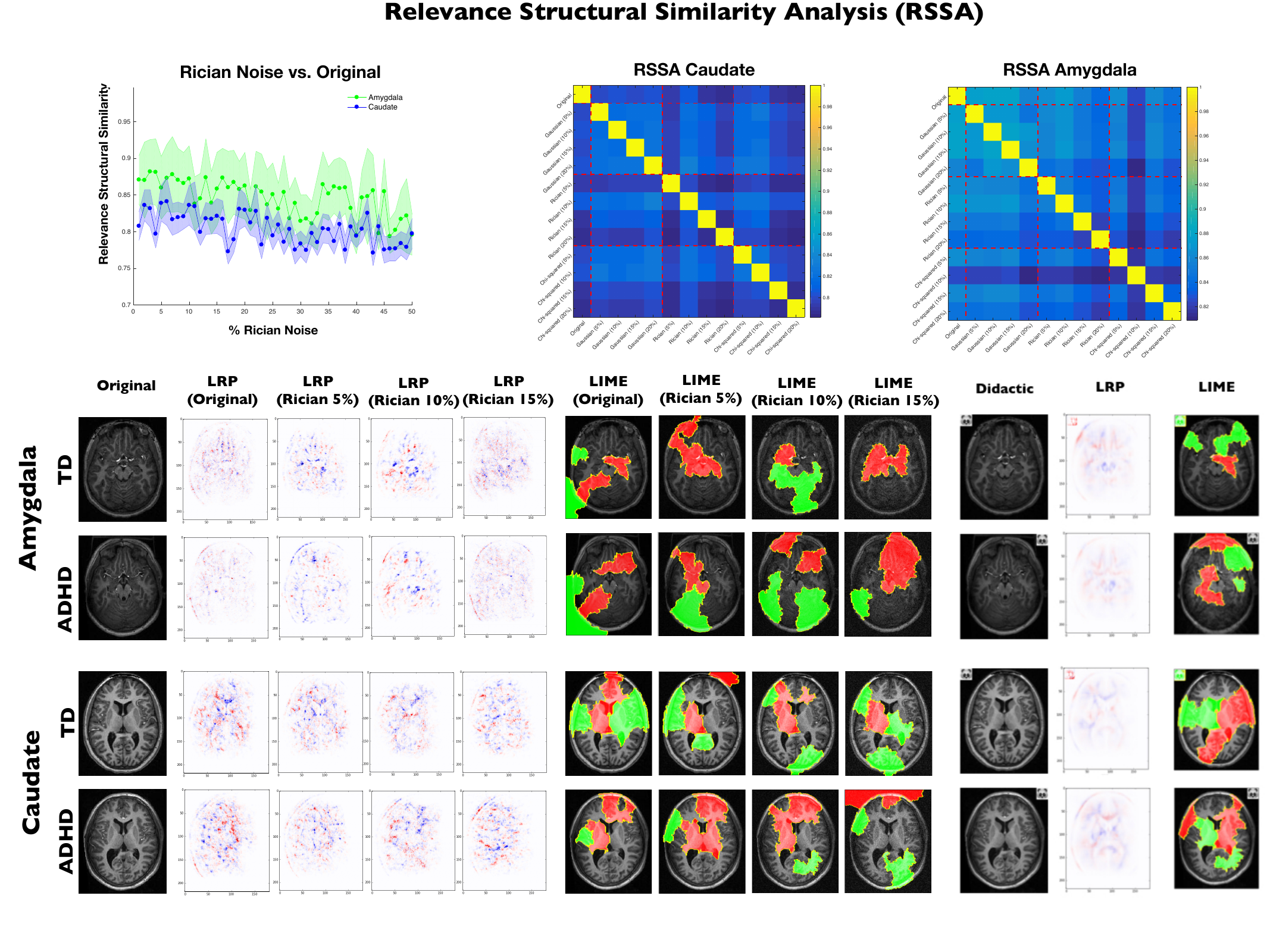}
    \caption{Relevance Structural Similarity Analysis (RSSA) results. (Top Left) RSS across Rician noise levels. (Top Middle, Right) RSSA matrix across each noise type compared to original data. (Bottom) RSSA maps for LIME and LRP for Rician noise and didactic examples.}
    \label{fig:fig2.jpeg}
\end{figure}

\section{Conclusions}

Explainable deep learning is crucial for medical applications where results may inform treatment and triage. Few studies have examined the extent to which data augmentation techniques alter credit assignment. Classifier performance varied with the frequency of adversarial augmentation according to a ‘w-curve', and RSSA revealed LRP to be more stable than LIME. Positive and negative controls are ubiquitiously appled in biology for benchmarking purposes.  Similarly, a ground truth for explainable deep learning is imperative.  Harnessing didactic examples may prove useful for validating deep learning explanations.  

\section{Acknowledgments}
We gratefully acknowledge the Air Force Office of Scientific Research (AFOSR) for grant FA9550-20-1-0042, which supported this work. 

\section*{References}

Abolghasemi, P. Et al. “Pay Attention! - Robustifying a Deep Visuomotor Policy Through Task-Focused Visual Attention” The IEEE Conference on Computer Vision and Pattern Recognition (CVPR), 2019, pp. 4254-4262

Bach, S. Et al. “Controlling Explanatory Heatmap Resolution and Semantics via Decomposition Depth,” ICIP (2016)

Bach et al. On Pixel-Wise Explanations for Non-Linear Classifier Decisions by Layer-Wise Relevance Propagation,, PLoS ONE 2015 

Böhle, M. Et al. “Layer-Wise Relevance Propagation for Explaining Deep Neural Network Decisions in MRI-Based Alzheimer's Disease Classification.” Front Aging Neuroscience 11:194 (2019)

Douglas, P. K. et al.: Single trial decoding of belief decision making from EEG and fMRI data using independent components features. Frontiers in Human Neuroscience \textbf{7}, (2013).

Douglas, PK, “Hemispheric brain asymmetry differences in youths with attention-deficit/hyperactivity disorder” Neuroimage: Clinical 18, 744-752. (2018). 

Evtimov, I et al. “Robust physical-world attacks on deep learning models,” arXiv:1707.08945, 2017.

Gastaldi, X. Shake-shake regularization. ICLR Workshop, 2017.

Hoogman, M et al. Subcortical brain volume differences in participants with attention deficit hyperactivity disorder in children and adults: a cross-sectional mega-analysis. The Lancet Psychiatry 4(4) 310-319. (2017)

Kriegeskorte, N. and Douglas, P. K.: Cognitive Computational Neuroscience. Nature Neuroscience. \textbf{21}. 1148-1160 (2018).

Khaligh-Razavi, S.-M., and Kriegeskorte, N.: Deep supervised, but not unsupervised, models may explain IT cortical representation. PLOS Computational Biol. 10 (2014).

Kriegeskorte, N. and Douglas, P. K.: Interpreting Encoding and Decoding Models. arXiv:1812.00278 [q-bio] (2018).

Lapuschkin et al., Analyzing Classifiers: Fisher Vectors and Deep Neural Networks, CVPR 2016 

Ribeiro, MT, Singh, S, Guestrin, C. “Why Should I Trust You?” Explaining the Predictions of Any Classifier. ACM KDD (2016)

Zeiler, MD, Fergus, R. “Visualizing and understanding convolutional networks,” in ECCV, 2014, pp. 818–833. 

Zhou, W., A. C. Bovik, H. R. Sheikh, and E. P. Simoncelli. "Image Qualifty Assessment: From Error Visibility to Structural Similarity." IEEE Transactions on Image Processing. Vol. 13, Issue 4, April 2004, pp. 600–612.

Zou, J. Et al. “A primer on deep learning in genomics” Nature Genetics. volume 51, pages 12–18 (2019)

\end{document}